\documentclass[sigconf]{acmart}

\usepackage{tikz}
\usepackage{standalone}

\usepackage{amsthm,amsmath}
\usepackage{dsfont}
\usepackage{booktabs}
\usepackage{xcolor}
\usepackage{graphicx}
\usepackage{subcaption}
\usepackage{tablefootnote}
\usepackage{bbding}
\usepackage{pifont}
\usepackage{wasysym}
\usepackage{wrapfig}
\AtBeginDocument{%
  }

\setcopyright{acmlicensed}
\copyrightyear{2025}
\acmYear{2025}
\acmDOI{XXXXXXX.XXXXXXX}
\acmConference[TGL Workshop, KDD 2025]{Temporal Graph Learning Workshop, SIGKDD International Conference on Knowledge Discovery and Data Mining 2025}{August 03-07,
  2025}{Toronto, Canada}

\acmISBN{978-1-4503-XXXX-X/2018/06}

\usepackage{multirow}

\begin{document}


\title[Transferring Insights from Evaluating Recommender Systems to Temporal Link Prediction]{Are We Really Measuring Progress? Transferring Insights from Evaluating Recommender Systems to Temporal Link Prediction}

\author{Filip Cornell}
\authornote{Both authors contributed equally to this research.}
\email{filip.cornell@king.com}
\affiliation{%
  \institution{King/Microsoft}
  \city{Stockholm}
  \country{Sweden}
}
\author{Oleg Smirnov}
\authornotemark[1]
\authornote{Corresponding author.}
\email{oleg.smirnov@microsoft.com}
\affiliation{%
  \institution{King/Microsoft}
  \city{Stockholm}
  \country{Sweden}
}

\author{Gabriela Zarzar Gandler}
\email{gabriela.zarzar@king.com}
\affiliation{%
  \institution{King/Microsoft}
  \city{Stockholm}
  \country{Sweden}}

\author{Lele Cao}
\email{lele.cao@king.com}
\affiliation{%
  \institution{King/Microsoft}
  \city{Stockholm}
  \country{Sweden}
}

\renewcommand{\shortauthors}{Cornell, Smirnov, et al.}

\begin{abstract}
    Recent work has questioned the reliability of graph learning benchmarks, citing concerns around task design, methodological rigor, and data suitability. In this extended abstract, we contribute to this discussion by focusing on evaluation strategies in Temporal Link Prediction (TLP). We observe that current evaluation protocols are often affected by one or more of the following issues: (1) inconsistent sampled metrics, (2) reliance on hard negative sampling often introduced as a means to improve robustness, and (3) metrics that implicitly assume equal base probabilities across source nodes by combining predictions. We support these claims through illustrative examples and connections to longstanding concerns in the recommender systems community. Our ongoing work aims to systematically characterize these problems and explore alternatives that can lead to more robust and interpretable evaluation. We conclude with a discussion of potential directions for improving the reliability of TLP benchmarks.
\end{abstract}

%
%




\settopmatter{printfolios=true}

\maketitle

\section{Introduction}

\begin{table}[t]
\centering
\resizebox{\linewidth}{!}{%
\begin{tabular}{@{}lccc@{}}
\textbf{Benchmark} & 
\rotatebox{45}{\shortstack{Inconsistent\\Sampled Metrics}} & 
\rotatebox{45}{\shortstack{Hard Negative\\Sampling}} & 
\rotatebox{45}{\shortstack{Combined Nodes \\ Predictions}} \\ \midrule
DGB~\citep{poursafaei2022towards}          & \textcolor{gray}{\ding{110}} & \textcolor{gray}{\ding{110}} & \textcolor{gray}{\ding{110}} \\
TGB~\citep{huang2024temporal}             & \textcolor{gray}{\ding{110}} & \textcolor{gray}{\ding{110}} & \textcolor{gray}{\ding{111}} \\
BenchTemp~\citep{huang2024benchtemp}      & \textcolor{gray}{\ding{110}} & \textcolor{gray}{\ding{111}} & \textcolor{gray}{\ding{110}} \\
TGB-Seq~\citep{yi2025tgbseq}              & \textcolor{gray}{\ding{110}} & \textcolor{gray}{\ding{111}} & \textcolor{gray}{\ding{111}} \\
DyGLib~\citep{yu2023towards}             & \textcolor{gray}{\ding{110}} / \textcolor{gray}{\ding{111}} & \textcolor{gray}{\ding{110}} & \textcolor{gray}{\ding{110}} \\ 
\bottomrule
\end{tabular}%
}
\caption{
Evaluation issues across TLP benchmarks and libraries. \textcolor{gray}{\ding{110}} indicates the issue is present; \textcolor{gray}{\ding{111}} indicates absence. 
For DyGLib, the authors additionally report the uniformly sampled average rank, which avoids the inconsistency issue.
}
\label{tab:issues-identified}
\vspace{-20pt}
\end{table}

With the increasing interest in learning over graphs, Temporal Graph Learning (TGL) has gained significant traction, leading to the development of numerous neural architectures. A variety of Temporal Graph Neural Networks (TGNNs) have been proposed~\citep{wang2021inductive, rossi2020temporal, kumar2019predicting, cong2023we, wang2021tcl, gravina2024long, Xu2020Inductive, yu2023towards, ding2024dygmamba}, alongside several distinct evaluation protocols~\citep{poursafaei2023exhaustive, huang2024temporal, poursafaei2022towards}. Notably, under these evaluation schemes, prior studies have demonstrated that simple heuristic methods can outperform neural models in both temporal graph and Temporal Knowledge Graph tasks~\citep{cornell2025on, dileo2024link, gastinger2024history}. 

This observation raises concerns about the complexity of existing datasets. \citet{yi2025tgbseq} and \citet{gastinger2024history} suggest that repetitive structures in datasets may lead to predictable patterns. If such patterns dominate, learning-based methods should ideally capture and build upon them more effectively.

Another key limitation of many TGL models is their poor scalability to large datasets. These models typically maintain time-dependent node embeddings that evolve at each timestamp, making standard retrieval methods, such as kNN-based lookup, impractical. To manage this, prior work has adopted sampling-based evaluation strategies to enable experimentation on large-scale workloads.

A recent position paper~\citep{bechler2025position} calls for a substantial revision of benchmarks in the broader graph learning field. Among the key concerns is the choice of evaluation strategy, which fundamentally influences how model performance is interpreted and compared.

This work contributes to that discussion by examining evaluation strategies used in TLP. We illustrate how specific metric choices may lead to inconsistent or misleading conclusions, which we illustrate through practical examples. Our analysis draws on prior work~\citep{krichene2020sampled, poursafaei2023exhaustive}, particularly from the recommender systems domain, where similar issues have long been recognized. We further present empirical results on TLP benchmarks that reveal critical flaws in current evaluation protocols. These findings further motivate the need for continued scrutiny in benchmark and metric design, aligning with concerns raised in prior work.





\section{Issues in TLP evaluation} \label{section:eval-issues}
This section reviews evaluation strategies in TLP, highlighting how they can produce misleading impressions of model performance. We illustrate these issues through prior work in recommender systems and with practical and illustrative examples.

We examine five established benchmarks and libraries: Dynamic Graph Benchmark (DGB)~\citep{poursafaei2022towards}, Temporal Graph Benchmark (TGB)~\citep{huang2024temporal}, BenchTemp~\citep{huang2024benchtemp}, DyGLib~\citep{yu2023towards}, and TGB-Seq~\citep{yi2025tgbseq}. These, along with most TLP studies, adopt ROC-AUC and Average Precision (AP) for binary classification~\citep{li2023towards, huang2024benchtemp, yu2023towards}, or sampled Mean Reciprocal Rank (MRR) for ranking tasks~\citep{yi2025tgbseq, huang2024temporal}.

\citet{rahman2025rethinking} frame these paradigms as questions: binary classification asks \textit{``Does an edge exist between nodes $u$ and $v$ at time $t$?''}, while ranking asks \textit{``Which nodes are likely to connect with node $u$ at time $t$?''}

Both formulations exhibit limitations that merit closer examination. Table~\ref{tab:issues-identified} summarizes the presence of three key concerns, across these benchmarks, discussed in details in following sections: (i) inconsistent sampled metrics, (ii) hard negative sampling, and (iii) combining predictions across nodes.

\subsection{Inconsistent Sampled Metrics} \label{sec:ics}
Numerous works~\citep{zhao2022revisiting, krichene2020sampled, li2020sampling, li2023towards, jin2021estimating, li2024item, liu2023consistency, liu2023we, dallmann2021case} in the recommender systems domain have examined the inconsistencies of top-$k$ ranking metrics under sampling. \citet{krichene2020sampled} analytically demonstrate that such metrics are not consistent under uniform sampling, even in expectation. They identify AUC as the only truly consistent metric due to its linearity, and propose three rank estimators to mitigate inconsistency: an unbiased estimator, a minimal bias estimator, and a variance-reducing extension of the latter.

Concurrently, \citet{canamares2020target} analyzed target set effects and reached similar conclusions. \citet{hidasi2023widespread} argue that sampled ranking metrics represent a pervasive flaw in the field. \citet{li2020sampling} and subsequent works~\citep{jin2021estimating, li2023towards, li2024item} introduce a series of correction techniques aimed at approximating true top-$k$ metrics from sampled data. More recently, \citet{liu2023consistency} explored how the robustness, consistency, and discriminative power of ranking metrics depend on dataset properties. \citet{liu2023we} later proposed a sampling strategy that adapts based on the number of items each user has interacted with.



\textbf{Sampled ranking metrics in TLP.} Two prominent benchmarks, TGB and TGB-Seq, employ ranking metrics in the form of sampled MRR. This is equivalent to the simplified sampled AP analyzed by \citet{krichene2020sampled}, who showed that such metrics are inconsistent under sampling. Sampling is used because full-ranking evaluations are computationally expensive and do not scale well for many current TGNNs~\citep{huang2024temporal, yi2025tgbseq}. To investigate alignment issues between full and sampled ranking metrics, we require evaluations in both settings. To this end, we employ scalable heuristics shown to be effective on TGB and BenchTemp~\citep{cornell2025on}. These heuristics score destination nodes based on either timestamp (recency) or occurrence count (popularity), operating at a local scale (per source node) or a global scale (entire dataset).

\begin{figure}
    \centering
    \includegraphics[width=.8\linewidth]{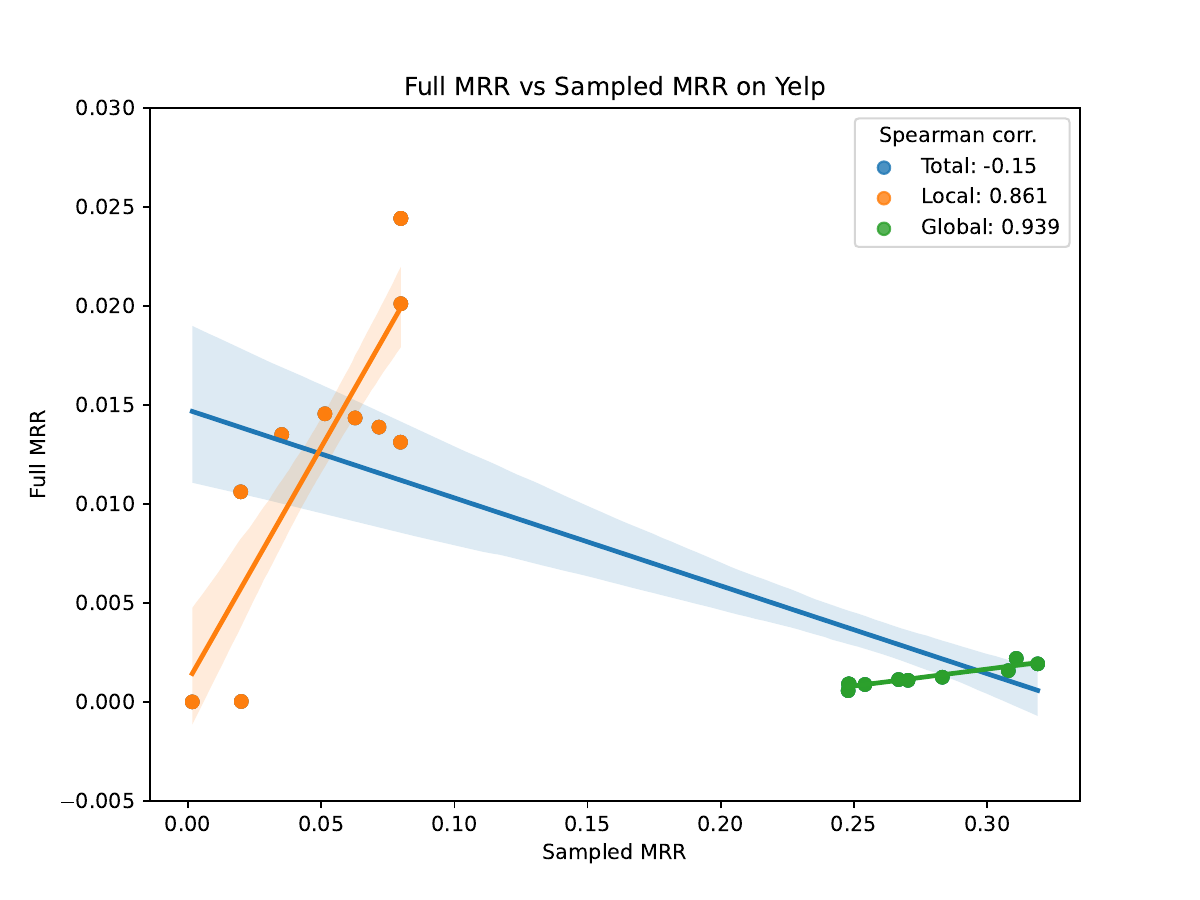}
    \caption{Sampled vs. full MRR for local and global heuristics, highlighting inconsistency across evaluation scales.}
    \label{fig:simpsons-paradox}
\end{figure}

Using these heuristics, we observe a Simpson's paradox~\citep{hernan2011simpson} on certain datasets. Figure~\ref{fig:simpsons-paradox} illustrates this phenomenon using the Yelp dataset from TGB-Seq~\citep{yi2025tgbseq}. Heuristics with \textit{local} granularity, which track destination scores per source node, show strong internal positive correlation. However, when combined with \textit{global} heuristics, the overall correlation turns strongly negative. Conversely, excluding the local heuristics, the global ones remain positively correlated among themselves.

This suggests that small changes in inductive biases~\citep{mitchell1980need} of a model yield consistent results within similar heuristic types, while more significant differences introduce greater inconsistency. While not rigorously tested on conventional TGNN models due to the prohibitive computational cost of running inference for each node at every timestamp in any real-world settings, we hypothesize that similar effects may also manifest in these models as well. In particular, the fact that algorithmic heuristics yield sampled metric scores closely aligned with those of TGNNs (ref. Table~2 in~\citep{cornell2025on}) suggests that correlation risks extending to other evaluation protocols.

As part of our ongoing work, we explore using covariance estimates between model predictions and negative samplers to assess bias in sampled ranking metrics. While challenging in practice due to non-uniform covariation and skewed score distributions, this approach may help identify when model comparisons are unfair and guide the development of principled correction methods.

\subsection{Hard Negative Sampling} \label{sec:hardnegs}


Sampling harder negatives for improved evaluation in TLP was first proposed by the authors of DGB~\citep{poursafaei2022towards}, who observed that the binary classification task often becomes too easy. Under uniform sampling, evaluation metrics can yield near-perfect scores, a pattern also evident in the BenchTemp leaderboard.

\textbf{Negative sampling strategies in TLP evaluation.} While hard negative sampling is often introduced to improve evaluation robustness, it is well established in the recommender systems community~\citep{zhao2022revisiting, hidasi2023widespread, dallmann2021case, liu2023consistency} that it can lead to inconsistent conclusions. Its impact varies across models and workloads, and although it may improve robustness in specific cases, it is not considered a generally reliable evaluation strategy. For example, \citet{dallmann2021case} show that popularity-based sampling, similarly to uniform sampling, can produce results inconsistent with full-ranking metrics. \citet{hidasi2023widespread} also discuss candidate generation and note that although better candidate generation methods may mitigate the issue, no universally accepted solution currently exists.

While a general drop in model performance is expected under hard negative sampling as noted in prior work~\citep{poursafaei2022towards}, the key question is: \textit{given a specific sampler, does each model's metric converge to its true value at the same rate compared to uniform sampling?} If not, model rankings may shift, leading to potential underestimation or overestimation of performance metrics for some models. In certain settings, performance estimates are observed to be less reliable than those from uniform sampling, as illustrated by the \texttt{tgbl-review} dataset. We discuss this issue in further details in Appendix~\ref{appx:tgbl-review}.

Ranking discrepancies of this kind have already been highlighted in the TLP domain~\citep{poursafaei2023exhaustive, poursafaei2022towards, yu2023towards}. For example, based on results from~\citet{poursafaei2022strong}, we computed Spearman rank correlations of the average test AUCs of all models reported under different negative sampling strategies, as shown in Table~\ref{tab:dgb-spearman-corrs}. In at least one dataset for each of the three sampling strategies, the correlation is \textit{negative}, indicating inconsistent or even reversed model rankings. Across all measurements, only 8 out of 39 correlations approach 0.9, underscoring the high variability in comparative model performance. Computing analogous statistics for nine models using the setup from \citet{yu2023towards} yields the same pattern.

As several works have pointed out~\citep{poursafaei2022towards, poursafaei2023exhaustive, yu2023towards}, different biased evaluations capture different aspects of model behavior. While this enables valuable analysis, it raises the fundamental question: \textit{which sampled metric should be trusted when selecting a model?} Choosing the maximum result across samplers risks aligning with aspects not reflected in real deployment, leading to overestimation. Without alignment between the sampling strategy and the inference-time setup, real-world performance estimates may become difficult to interpret reliably. This concern has been raised in prior work, notably by~\citet{hidasi2023widespread} and~\citet{castells2022offline}, the latter arguing that the ideal candidate selector in offline evaluation should mirror the one used in production systems.

Our ongoing work investigates a debiasing strategy for hard negative sampling that aggregates multiple biased samplers. By combining diverse candidate sources, the approach aims to mitigate individual sampler biases and approximate a more balanced evaluation signal.

\subsection{Combined Nodes Predictions} \label{sec:ebp}
The final issue we identify in TLP evaluation arises when edge predictions from different source nodes are combined. 
This approach implicitly assumes equal base probabilities for all edges, regardless of the query (source) node. A similar concern was raised in the recommender systems domain by~\citet{tamm2021quality}, who questioned the validity of aggregating recommendations across users. They proposed computing AUC on a per-user basis as a more appropriate alternative, though without fully addressing its limitations. This per-user AUC formulation is adopted in OpenRec~\citep{yang2018openrec} and used by \citet{zhu2017optimized}.

\begin{figure}
    \centering
    \includegraphics[width=0.8\linewidth]{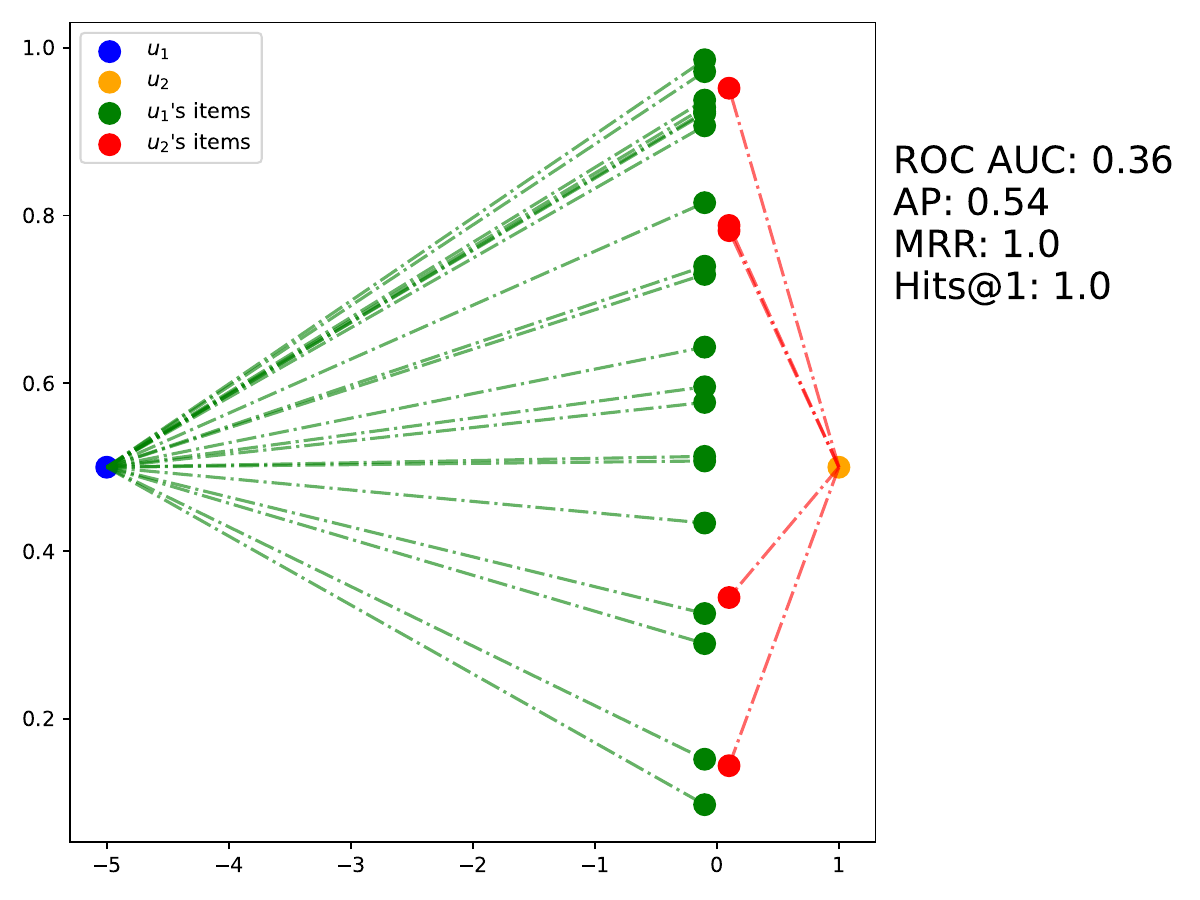}
    \caption{An example where ranking metrics computed per source node yield perfect scores, while aggregated metrics such as AUC and AP report substantially lower values. }
    \label{fig:uniformity}
\end{figure}

To illustrate this issue, consider the example in Figure~\ref{fig:uniformity}, showing two source nodes, $U_1$ and $U_2$, with different distances to their respective targets. The filtered ranking metrics yield perfect scores, as all queries $Q = \{(u_1, v_i) \in E\} \cup \{(u_2, v_i) \in E\}$ receive a filtered rank of 1. However, both ROC AUC and AP report much lower values. This occurs because the negative samples associated with $U_1$ are ranked above the true edges of $U_2$, resulting in poor rankings despite perfect per-query results.

The problem intensifies when a source node appears frequently in the dataset. Such a node, having a higher base probability of forming edges, may rank closer even to incorrect targets, causing the true edges of less frequent nodes to be ranked lower. In graphs with power law degree distributions, hub nodes can dominate the sampled evaluation, leading to misleading metric values.

How significant is this issue? From a graph-theoretic perspective, node degrees in real-world graphs often follow a power-law distribution~\citep{barabasi1999emergence}, where a small number of nodes have very high degree while the majority have low degree. As a result, high-degree nodes inherently have a higher base probability of forming edges and are more likely to appear in the test set. We conjecture that this imbalance can substantially affect evaluation outcomes, particularly as a function of source node degree distribution in the evaluation set. 

\begin{table}[]
    \centering
    \resizebox{.75\linewidth}{!}{%
    \begin{tabular}{llll}
\toprule
\multirow{2}{*}{Sampling strategy} & Historical & Historical & Inductive \\
 & Inductive & Random & Random \\
\midrule
Can. Parl. & 0.99 & 0.38 & 0.36 \\
Contact & 0.34 & 0.16 & 0.02 \\
Enron & 0.49 & 0.2 & -0.27 \\
Flights & -0.1 & -0.22 & 0.52 \\
LastFM & 0.29 & -0.11 & -0.23 \\
MOOC & 0.85 & 0.94 & 0.93 \\
Reddit & 0.75 & 0.84 & 0.9 \\
Social Evo. & 0.93 & 0.89 & 0.89 \\
UCI & 0.71 & 0.63 & 0.7 \\
UN Trade & 0.11 & 0.52 & 0.16 \\
UN Vote & 0.64 & 0.41 & 0.77 \\
US Legis. & 0.83 & 0.54 & 0.21 \\
Wikipedia & 0.69 & 0.82 & 0.95 \\
\bottomrule
\end{tabular}%
}
    \caption{Spearman rank correlations of seven models based on their AUC rankings, using \citet{poursafaei2022strong} results.}
    \label{tab:dgb-spearman-corrs}
\end{table}

\section{Remedies and solutions}
Multiple solutions have been proposed in prior work, and here we outline a few promising directions in the context of TLP.


\textbf{Avoiding sampling schemes completely.} Several works~\citep{krichene2020sampled, zhao2022revisiting, hidasi2023widespread} recommend avoiding sampled top-$k$ ranking metrics entirely. One promising direction is to develop model architectures that support full-scale evaluation or kNN-based retrieval during inference, enabling accurate top-$k$ metrics. Recent TLP benchmarks have made progress in this direction: both TGB and TGB-Seq report training and inference runtimes, while BenchTemp incorporates runtime directly into its leaderboard metrics.

Pushing further remains valuable. A compelling example is the WikiKG90Mv2 benchmark from the OGB-LSC competition~\citep{hu2021ogblsc}, where models must retrieve the top 10 entities from all 91 million entities, resulting in measuring $MRR@10$. This setup enforces scalability by design and highlights the importance of evaluation constraints aligned with deployment realities.


\textbf{Using consistent metrics.} If sampled metrics are still required, more reliable alternatives exist. The simplified AUC analyzed by \citet{krichene2020sampled} corresponds to a consistent, normalized inverse estimate of the mean rank, denoted $\widehat{MR}$. Closely related is the hit ratio, which \citet{li2020sampling} show is linearly related to the expected sampled $\mathrm{Hits@}C \cdot K$ and $\mathrm{Hits@}K$, where $C = \frac{|V|-1}{n_s}$. 

These properties make $\widehat{MR}$ and $\mathrm{Hits@}C \cdot K$ consistent even when sampling a large fixed set of nodes for all queries, assuming appropriate handling of true answer collisions. This approach reduces the sampling complexity from $\mathcal{O}(|E_\text{test}|)$ to $\mathcal{O}(1)$, while maintaining statistical reliability. However, it is important to note that AUC captures different performance aspects compared to commonly used top-$k$ metrics in TLP~\citep{zhao2022revisiting}. Some recent works~\citep{yu2023towards, ding2024dygmamba} already report uniformly sampled mean rank, thereby adopting a more consistent evaluation.

It is important to emphasize that $\widehat{MR}$ assumes \textit{uniform} negative sampling. When harder negatives are used, the estimate becomes biased. Furthermore, $\widehat{MR}$ is not without limitations. If other true targets are filtered out during evaluation, this can affect the metric's reliability. As a result, we are restricted to using the \textit{raw} $\widehat{MR}$, where all unranked entities, including true ones, are treated as negatives. Under suitable conditions, e.g., when the number of filtered entities is small and the number of nodes is large, this bias can be considered negligible. However, if many true entities are filtered for a given query, the impact may become substantial.



\textbf{Using statistical corrected estimates.} Several works~\citep{krichene2020sampled,li2020sampling,li2024item,li2023towards,liu2023we} propose statistical correction techniques to improve the estimation of true ranking metrics. While often mathematically involved, these methods can reduce bias, though at the cost of increased variance. Their correction effectiveness has been questioned~\citep{zhao2022revisiting}, and further research is needed to adapt these approaches to TLP, where temporal dynamics must be taken into account.

\textbf{Using alternative evaluation protocols.} Some recent works have proposed alternative evaluation strategies to address limitations in existing metrics. For example, \citet{rahman2025rethinking} suggest a counterfactual evaluation method that perturbs timestamps. However, this approach still relies on metrics such as AP and ROC-AUC, which remain sensitive to inconsistent sampling and combines predictions across source nodes.

Earlier, \citet{poursafaei2023exhaustive} introduced EXHaustive, which avoids sampling by partitioning edges into time intervals, thereby reducing repeated evaluations. This strategy highlights performance drops and ranking inconsistencies also reported by~\citet{yu2023towards}. However, EXHaustive has practical limitations and may introduce its own biases. It requires domain knowledge, does not scale well with dense temporal datasets, and also introduces its own biases. In particular, it assumes that the model should prioritize edges selected by the evaluation strategy itself -- an assumption that may not hold for many models.

While these methods represent useful steps forward, they may not fully address the underlying challenges and can shift the bias rather than remove it. As a result, their limitations need to be carefully considered before drawing conclusions.

\section{Conclusion}

This work highlights the need for a unified and reliable evaluation framework for TLP, building on insights extensively discussed in the recommender systems literature. Ideally, such a framework should be easy to use in practice and enable confident interpretation of model performance. However, the assumptions underlying current evaluation protocols often undermine this goal. These issues may not appear in every setting, as their impact depends on interactions between dataset characteristics, model behavior, and evaluation design. Still, even isolated counterexamples can demonstrate how current methods may yield unreliable or misleading results if their limitations are not well understood.

In the absence of a perfect solution, our suggestions are as follows. First, models should be designed with reliable evaluation in mind. Second, if sampling is used, it should yield statistically unbiased estimates or be paired with a suitable correction method. Third, biased negative sampling should be avoided when possible; if used, conclusions about model performance should be limited to scenarios where the sampling strategy aligns with the inference-time setting.

Looking ahead, our ongoing work aims to address these limitations through a covariance-based correction method for sampling bias and a multi-sampler strategy to mitigate distortions from hard negative sampling.  We believe these directions can contribute to more robust and interpretable evaluation, and support the development of fair, reproducible benchmarks. More broadly, continued exchange between TGL and recommender systems domains may inspire more principled evaluation practices.

\begin{acks}
This work was partially supported by Wallenberg AI, Autonomous Systems and Software Program (WASP). 
\end{acks}

\bibliographystyle{ACM-Reference-Format}
\bibliography{ref}

\appendix

\section{Exemplifying the Issue of Biased Negative Sampling}\label{appx:tgbl-review}

To illustrate the effects of biased negative sampling, we draw on an observation from~\citet{cornell2025on}. While many TGB link prediction benchmarks contain a high degree of edge repetition, \texttt{tgbl-review}, which consists of Amazon product reviews, stands out with a surprise index of 98.7\%. This means that only 1.3\% of historical edges reoccur. Despite this, historical edges are still used as negative samples.

In this setting, simple heuristics can rule out a large portion of negative samples by assigning low scores to edges that have appeared in the past. Machine learning models, such as GNNs and other embedding-based approaches, may however still assign high scores to these edges due to similarities in their learned representations. As a result, models can be penalized for favoring edges that heuristics would easily dismiss. On the other hand, a model that learns to suppress historical edges may achieve artificially high scores, producing an overly optimistic estimate when compared to uniform sampling of a similar size.

This example highlights that alignment between a model’s inductive bias and the sampling strategy may play a more influential role than the alignment between the sampler and the dataset itself.

\end{document}